\documentclass[10pt,twocolumn,letterpaper]{article}
\PassOptionsToPackage{nameinlink}{cleveref}
\PassOptionsToPackage{dvipsnames}{xcolor}
\usepackage[pagenumbers]{cvpr}
\usepackage[T1]{fontenc}
\usepackage{comment}
\usepackage{listings}
\usepackage{calc}
\usepackage{adjustbox}
\usepackage[percent]{overpic}
\input{macros}

\usepgfplotslibrary{groupplots,dateplot}
\usetikzlibrary{patterns,shapes.arrows,decorations.pathreplacing,calc,automata,shapes.geometric,decorations.pathmorphing,matrix,chains,scopes,positioning,arrows,fit}

\definecolor{cvprblue}{rgb}{0.21,0.49,0.74}
\usepackage[breaklinks,colorlinks,allcolors=cvprblue]{hyperref}
\hypersetup{
  pdftitle={point2vec for Self-Supervised Representation Learning on Point Clouds},
  pdfauthor={Karim Knaebel, Jonas Schult, Alexander Hermans, Bastian Leibe}
}

\AtEndPreamble{
  \crefname{pseudocode}{Pseudocode}{Pseudocodes}
  \Crefname{pseudocode}{Pseudocode}{Pseudocodes}
}

\title{point2vec for Self-Supervised Representation Learning on Point Clouds}
\author{
  Karim Knaebel$^*$ \qquad Jonas Schult$^*$ \qquad Alexander Hermans \qquad Bastian Leibe\\[0.5em]
  RWTH Aachen University\\[0.5em]
  {\ttfamily \href{https://vision.rwth-aachen.de/point2vec}{https://vision.rwth-aachen.de/point2vec}}
}

\begin{document}
\maketitle
\begingroup
\renewcommand{\thefootnote}{*}
\begin{NoHyper}
\footnotetext[1]{Equal contribution.}
\end{NoHyper}
\endgroup

\begin{abstract}
We present \name{}, a self-supervised method for learning point cloud representations through masked prediction of a teacher network's latent features.
Building on data2vec, we investigate a limitation of applying this approach to point clouds: positional information associated with masked patches reveals the overall object shape to the student, weakening the learning task even under heavy masking.
\name{} addresses this issue by passing only visible patches to the student encoder and introducing a shallow decoder to predict representations for masked patches.
The teacher processes the complete point cloud to provide contextualized training targets.
These targets capture global context through the teacher's self-attention, providing supervision beyond the local geometry of individual patches.
This design conceals masked-patch positions from the student and reduces its computational cost during pre-training.
Our ablations show that the decoder alone provides only part of the improvement; concealing positional information is key.
After pre-training on ShapeNet, \name{} outperforms the evaluated self-supervised baselines on shape classification on ModelNet40 and ScanObjectNN and few-shot classification on ModelNet40, while achieving competitive part segmentation results on ShapeNetPart.

\end{abstract}

\section{Introduction}
\begin{figure}[t]
\centering

\includegraphics[width=1.0\linewidth]{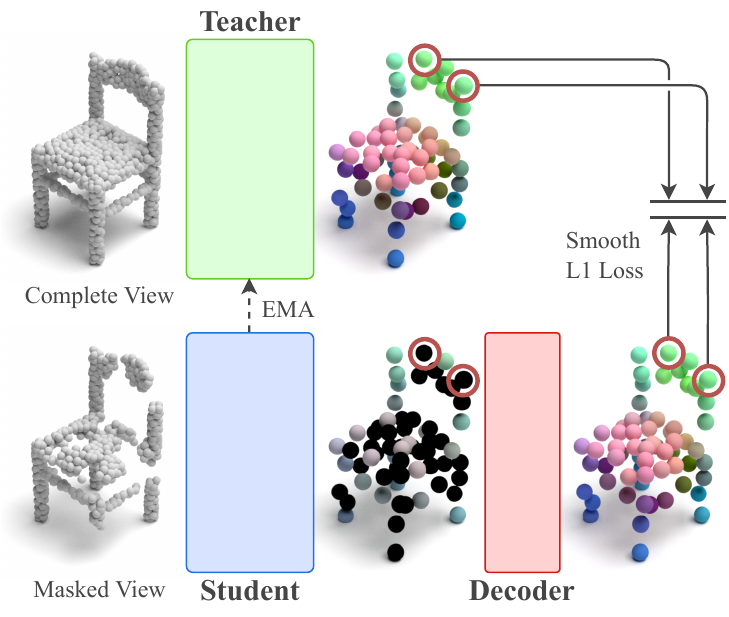}

\caption{\textbf{Overview of \name{}.}
During training, a teacher network~\colorsquare{m_green} predicts latent representations using a complete view of the point cloud.
The student network~\colorsquare{m_blue} encodes a partial view of the same point cloud.
A shallow decoder~\colorsquare{m_red} predicts the teacher's representations for masked regions~\colordot{black}, providing the training signal for the student and decoder.
The teacher weights are updated using an exponential moving average of the student weights.
}
  \label{fig:teaser}
\end{figure}

In this work, we address the task of self-supervised representation learning on 3D point clouds.
With the ever-increasing availability of affordable consumer-grade 3D sensors, point clouds are becoming a widely adopted data representation for capturing real-world objects and environments~\cite{dai2017scannet, dehghan2021arkitscenes, Armeni16CVPR, behley2019semantickitti, caesar2020nuscenes}.
They provide accurate 3D geometric information, making them a valuable input for applications in robotics, autonomous driving~\cite{behley2019semantickitti, caesar2020nuscenes}, and AR/VR.
The 3D computer vision community has made impressive progress by developing 3D-centric approaches which directly process 3D point clouds to semantically understand 3D objects and environments~\cite{qi2016pointnet, qi2017pointnetplusplus, choy2019minkowskinet, schult23mask3d}.
However, these approaches typically rely on fully supervised training from scratch~\cite{xie2020pointcontrast}, requiring time-consuming and labor-intensive human annotations.
For example, semantically annotating a single room-scale scene of the ScanNet dataset takes about \SI{22}{\minute} on average~\cite{dai2017scannet}.
This results in a lack of large-scale annotated point cloud datasets, making it challenging to learn strong representations from limited data.

At the same time, self-supervised training has shown impressive results in natural language processing~\cite{zhilin2019xlnet,devlin2018bert}, speech~\cite{baevski2020wav2vec, hsu2021hubert}, and 2D vision~\cite{he2022mae,grill2020BYOL,baevski2022data2vec,caron2021dino,chen2020simclr}, enabling the learning of meaningful representations from massive unlabeled datasets without any human annotations.
Only recently, we have seen self-supervised methods being successfully applied to Transformer architectures for 2D vision~\cite{caron2021dino, baevski2022data2vec, he2022mae} and 3D point clouds~\cite{pang2022pointmae, yu2021pointbert, zhang2022pointm2ae}.
\citet{baevski2022data2vec} propose data2vec, a modality-agnostic self-supervised learning framework showing competitive performance in speech recognition, image classification, and natural language understanding.
data2vec uses a joint-embedding architecture~\cite{grill2020BYOL, baevski2022data2vec, caron2021dino} with a \emph{student} Transformer encoder and a \emph{teacher} network parameterized as the exponential moving average of the student weights.
Specifically, the teacher first predicts latent representations using an uncorrupted view of the input, which the student network then predicts from a masked view of the same input.

In this paper, our aim is to apply data2vec-like pre-training to point clouds. The key difference to top-performing approaches for point cloud representation learning such as Point-MAE~\cite{pang2022pointmae}, Point-M2AE~\cite{zhang2022pointm2ae}, and Point-BERT~\cite{yu2021pointbert} is the target representation.
The self-attention in the teacher Transformer encoder of data2vec generates contextualized feature targets that contain \emph{global} information about the entire input.
In contrast, Point-MAE and Point-M2AE explicitly reconstruct only \emph{local} point cloud patches, and Point-BERT is restricted to a fixed-size vocabulary of token representations.
To apply data2vec to point clouds, we use the same underlying 3D-specific Transformer model as Point-BERT and Point-MAE.
In experiments, we show that these modality-specific adaptations to data2vec already enable competitive performance compared to highly 3D-specific self-supervised approaches~\cite{yu2021pointbert, pang2022pointmae, zhang2022pointm2ae, liu2022maskpoint, wang2021occo}.
Encouraged by these promising results, we perform a subsequent analysis that reveals a crucial shortcoming specific to point clouds that restricts data2vec's representation learning capabilities:
data2vec uses masked embeddings in the student network which carry positional information.
Unlike images, text, and speech, the positional information in point clouds contains semantic meaning, namely 3D point locations (\cref{fig:early_leakage}).
Feeding masked embeddings with positional information into the student network therefore reveals the overall object shape to the student, which makes the masking operation far less effective, as also reported by \citet{pang2022pointmae} in the context of masked autoencoders for point clouds.
Based on this analysis, we propose \name{}, which effectively addresses the leakage of positional information to the student and thus unleashes the full potential of data2vec-like pre-training for point clouds.
To this end, we exclude masked embeddings from the student network.
This prevents the overall object shape from being revealed, while also decreasing the computational cost.
Instead, we introduce a shallow decoder which processes masked embeddings together with the student's outputs and which is trained to regress the representations of the teacher (\cref{fig:teaser}).

Evaluating the quality of the learned representations on downstream tasks is a crucial step for analyzing self-supervised methods.
Our experiments demonstrate that, after pre-training on the ShapeNet dataset~\cite{chang2015shapenet}, \name{} outperforms other self-supervised methods on both the ModelNet40~\cite{wu2015modelnet40} and ScanObjectNN~\cite{uy2019scanobjectnn} shape classification benchmarks.
Additionally, \name{} achieves state-of-the-art performance on few-shot classification on ModelNet40 and competitive results on part segmentation on ShapeNetPart~\cite{yi16siggraph}.
These findings suggest that the learned representations are strong and transferable, indicating that \name{} is a promising approach for self-supervised point cloud representation learning.

To summarize, our contributions are:
\begin{enumerate}
    \item We extend the seminal work data2vec to the point cloud domain.
    \item In our experiments, we discover a crucial shortcoming of data2vec that limits its representation learning capabilities for point clouds: masked embeddings leak positional information to the student, revealing the overall object shapes even under heavy masking.
    \item We propose \name{}, which unleashes the full potential of data2vec-like pre-training for self-supervised representation learning by addressing the aforementioned shortcoming.
    \name{} learns strong and transferable features in a self-supervised manner, outperforming self-supervised approaches on several downstream tasks.
\end{enumerate}

\section{Related Work}

\paragraph{Self-supervised learning.}
Recently, self-supervised learning has gained much attention due to its promise to learn meaningful data representations without any human annotations.
At the heart of self-supervised learning is the pretext task, offering a vast range of diverse options.
One such line of work investigates contrastive learning objectives~\cite{chen2020simclr, he2020moco, oord2018cpc, tian2020cmc, tian2020makes}, \ie maximizing feature similarity across multiple views of the same training sample while simultaneously minimizing the similarity to other training samples.
Contrastive learning approaches typically rely on a careful choice of data augmentations, negative sample mining, or large batch sizes~\cite{grill2020BYOL}.
Addressing these limitations, non-contrastive approaches~\cite{grill2020BYOL, chen2021simsiam, bardes2022vicreg, baevski2022data2vec, caron2021dino} follow a joint-embedding architecture, \ie two copies of the same network are trained to produce similar latent representations for two views of the identical input.
Among them and most important to our work is data2vec~\cite{baevski2022data2vec}, which relies on a teacher first generating targets by predicting latent representations using the complete view of the input and a student which predicts these targets using only a masked view of the same input.
Inspired by data2vec's flexibility across a wide range of modalities, in this paper, we seek to unlock the full potential of data2vec-like pre-training for point clouds by specifically taking the unique characteristics of point clouds into account.

\paragraph{Self-supervised learning on point clouds.}
The success of self-supervised learning in 2D vision~\cite{grill2020BYOL, caron2021dino, he2022mae, chen2020simclr, baevski2022data2vec,bao2021beit, bardes2022vicreg,oord2018cpc}, natural language processing~\cite{devlin2018bert, baevski2022data2vec}, and speech~\cite{baevski2020wav2vec, baevski2022data2vec} has inspired a number of recent works proposing self-supervised learning frameworks for point cloud understanding tasks.
Among them, contrastive self-supervised frameworks are typically deployed for room-scale pre-training.
The pioneering work of \citet{xie2020pointcontrast} contrasts corresponding 3D points from multiple partial views of a reconstructed static scene, showing impressive improvements when fine-tuned on several scene-level downstream tasks.
Building on this, \citet{hou2021contrastivescene} propose to leverage both point-level correspondences and spatial contexts of 3D scenes.
In contrast to room-scale pre-training, we see a line of work developing self-supervised methods tailored towards single-object understanding tasks~\cite{wang2021occo, huang2021strl, sharma2020ssl, yu2021pointbert, pang2022pointmae, zhang2022pointm2ae, xue2022ulip, liu2022maskpoint, eckart2021parae, rao2020pointglr}.
They typically use the inherent structure and geometry of 3D point clouds to learn meaningful representations, \eg by explicitly reconstructing point cloud patches using the Chamfer distance~\cite{zhang2022pointm2ae, pang2022pointmae}, discriminating masked points from noise~\cite{liu2022maskpoint}, or performing point cloud completion for occluded regions~\cite{wang2021occo}.
Another line of work additionally leverages multi-modal information to improve the latent representation of 3D point clouds, \ie incorporating knowledge from models on 2D images~\cite{dong2023act,xue2022ulip,zhang2022pointclip,zhang2023i2pmae} or text descriptions~\cite{xue2022ulip,zhang2022pointclip}.
The advances of the above methods are orthogonal to our approach \name{}, as it operates on point clouds only.
Most relevant to our work are Transformer-based self-supervised learning approaches on point clouds.
Due to the success of pre-trained Transformer architectures in various domains~\cite{devlin2018bert, baevski2022data2vec, he2022mae, caron2021dino}, we have recently seen a shift towards pre-training Transformer-based approaches for point clouds~\cite{liu2022maskpoint, yu2021pointbert, zhang2022pointm2ae}.
Among them, Point-BERT~\cite{yu2021pointbert} introduces a standard ViT-like~\cite{dosovitskiy2020vit} backbone to point clouds and
extends BERT~\cite{devlin2018bert} pre-training to point clouds.
Point-MAE~\cite{pang2022pointmae} and Point-M2AE~\cite{zhang2022pointm2ae} follow the masked autoencoder (MAE) approach proposed by \citet{he2022mae}.
In contrast to these methods, we do not explicitly reconstruct masked point cloud patches but predict contextualized targets in the latent feature space, circumventing the need to define sophisticated distance metrics to compare point cloud patches.

\begin{figure*}[t!]

\includegraphics[width=1.0\linewidth, trim={0 0.3cm 0 0.1cm}, clip]{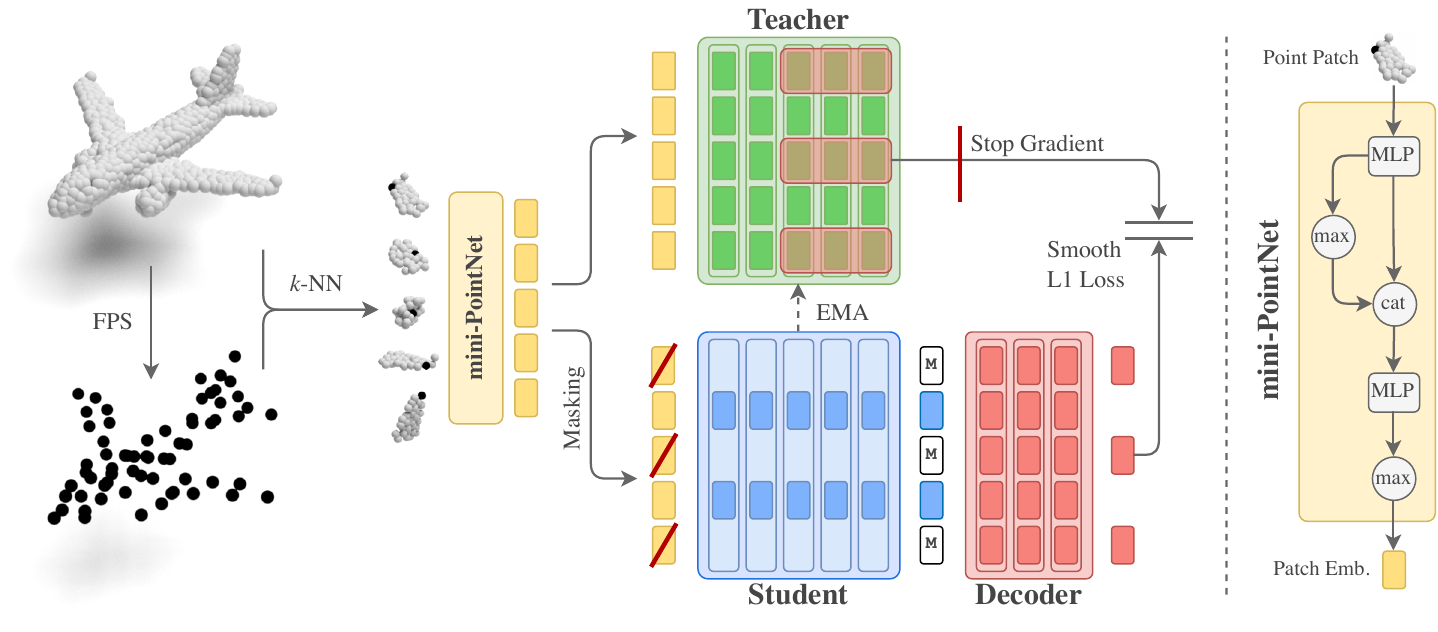}

\caption{
\textbf{point2vec pre-training.}
We divide the point cloud into patches using farthest point sampling (FPS) and $k$-NN aggregation.
We obtain patch embeddings by applying a mini-PointNet~\colorsquare{m_pointnet} to each point patch (right).
The teacher Transformer encoder~\colorsquare{m_green} infers contextualized representations for all patch embeddings, which, after normalization and averaging over the last $K$ Transformer layers, serve as training targets.
The student's input is a masked view of the input data, \ie we randomly mask out a fraction of patch embeddings and only pass the remaining embeddings into the student Transformer encoder~\colorsquare{m_blue}.
After applying a shallow decoder~\colorsquare{m_red} to the outputs of the student, padded with learned mask embeddings~\protect\maskembedding{}, we train the student and decoder to predict the teacher's latent patch representations.
}

\label{fig:model}
\end{figure*}

\section{Method}
The aim of this work is to unlock the full potential of data2vec-like~\cite{baevski2022data2vec} pre-training on point clouds by addressing challenges specific to point clouds.
To achieve this, we first summarize the technical concepts of data2vec (\cref{sec:method_d2v}) and show how to learn rich representations on point clouds using data2vec pre-training (\cref{sec:method_d2v_pcl}).
Finally, we propose \name{}, which accounts for the limitations of data2vec specific to point clouds (\cref{sec:method_p2v}).
\subsection{data2vec}\label{sec:method_d2v}
data2vec is designed to pre-train Transformer-based models, which involve a feature encoder that maps the input data to a sequence of embeddings.
These embeddings are subsequently passed to a standard Transformer encoder~\cite{vaswani2017attention} to generate the final latent representations.
During pre-training, two versions of the Transformer encoder are kept: a student and a teacher.
The teacher is a momentum encoder, \ie its parameters $\Delta$ track the student's parameters $\theta$ by being updated after each training step according to an exponential moving average (EMA) rule~\cite{caron2021dino, baevski2022data2vec, grill2020BYOL, he2020moco}:
\[
    \Delta \leftarrow \tau \Delta + (1-\tau)\theta,
\]
where $\tau \in [0,1]$ is the EMA decay rate.
The teacher provides the training targets, which the student predicts given a corrupted version of the same input.
In a first step, the teacher encodes the uncorrupted input sequence.
The training targets are then constructed by averaging the outputs of the last $K$ blocks of the teacher, which are normalized to prevent a single block from dominating the sum.
Due to the self-attention layers, these targets are contextualized, \ie they incorporate global information from the whole input sequence.
This is an important difference from other masked-prediction methods such as BERT~\cite{devlin2018bert} and MAE~\cite{he2022mae}, where the targets only comprise local information, \eg a word or an image patch. The student is given a masked version of the same input, where some of the embeddings in the input sequence are replaced by a special learned \emph{mask embedding}. The student's task is to predict the targets corresponding to the masked parts of the input.
The model is trained by optimizing a Smooth L1 loss on the regressed targets. 

\subsection{data2vec for Point Clouds}\label{sec:method_d2v_pcl}

To apply data2vec to point clouds, we utilize the same underlying model as Point-BERT~\cite{yu2021pointbert} and Point-MAE~\cite{pang2022pointmae}.
This model is well suited for data2vec pre-training: it extracts a sequence of patch embeddings from the input point cloud and feeds it to a standard Transformer encoder.
For downstream tasks, we append a task-specific head to the Transformer encoder (\cref{sec:experiments}).
Next, we describe the point cloud embedding and the Transformer in detail and conclude with a summary of data2vec for point clouds.

\paragraph{Point cloud embedding.}
First, we sample $n$ center points from the input point cloud using farthest point sampling (FPS)~\cite{qi2017pointnetplusplus}.
Grouping the center points' $k$-nearest neighbors ($k$-NN) in the point cloud yields $n$ contiguous \emph{point patches}, \ie sub-clouds of $k$ elements.
Next, we normalize the point patches by subtracting the corresponding center point from the patch's points.
This untangles the positional and the structural information.
To ensure invariance to the order of points,
we use a mini-PointNet~\cite{qi2016pointnet} (\cref{fig:model}, right) that maps each normalized point patch to a \emph{patch embedding}.

The mini-PointNet involves the following steps:
\begin{enumerate}
    \item Map each point of a patch to a feature vector using a shared MLP.
    \item Concatenate max-pooled features to each feature vector.
    \item Pass the resulting feature vectors through a second shared MLP and a final max-pooling layer to obtain the patch embedding.
\end{enumerate}

\paragraph{Transformer encoder.}
The central component of the model is a standard Transformer encoder.
The patch embeddings form the input sequence to the Transformer encoder.
Since the point patches are normalized, the patch embeddings carry no positional information;
therefore, a two-layer MLP maps each center point to a position embedding, which is then added to the corresponding patch embedding.
Due to the special importance of positional information in point clouds, the position embeddings are added again before each subsequent Transformer block to ensure that the positional information is incorporated at every step of the encoding process.

\paragraph{\datavec{}.}
Our baseline, \datavec{}, applies unmodified data2vec to the Point-BERT and Point-MAE backbone described above.

\subsection{\name{}}\label{sec:method_p2v}
In \cref{fig:model}, we present the complete pipeline of our \name{} model.
Directly applying data2vec to point cloud data without modifications is not optimal, as the position embeddings are also added to the mask embeddings, revealing the overall shape of the point cloud to the student.
As positions are the only features,
this makes the masking far less effective, as noted by \citet{pang2022pointmae} in the context of masked autoencoders (MAE).

To solve this issue, we adopt an approach inspired by MAE, where we only feed the non-masked embeddings to the student~\colorsquare{m_blue}.
A separate decoder~\colorsquare{m_red}, implemented as a shallow Transformer encoder, takes the output of the student and the previously held-back masked embeddings~\maskembedding{} as input and predicts the training targets.
In contrast to \datavec{}, this approach does not suffer from leaking positional information from the masked-out point patches to the student.
Moreover, utilizing an MAE-inspired setup provides additional benefits:
First, the student is more computationally efficient, as it only needs to process the non-masked embeddings.
Second, the model's inputs during fine-tuning are more similar to those during pre-training because
they are no longer dominated by masked embeddings, which are absent during fine-tuning.
This likely makes the learned representations more transferable to downstream tasks.

\section{Experiments}\label{sec:experiments}
In this section, we describe the self-supervised pre-training of \name{} on ShapeNet~\cite{chang2015shapenet} (\cref{sec:pretraining}).
Next, we compare \name{} with top-performing self-supervised approaches and our baseline method \datavec{} on three well-established datasets and four downstream tasks (\cref{sec:main_results}).
Finally, we put the spotlight on the architectural changes from our data2vec adaptation for point clouds to our proposed model \name{}, which address the unique challenges of 3D point clouds (\cref{sec:analysis}).
In the supplementary material, we provide detailed hyperparameters of our model.
Our code and checkpoints are publicly available.

\subsection{Self-Supervised Pre-Training}
\label{sec:pretraining}
Following the pre-training protocol used by Point-BERT~\cite{yu2021pointbert}, Point-MAE~\cite{pang2022pointmae}, and Point-M2AE~\cite{zhang2022pointm2ae}, we pre-train \name{} on the training split of ShapeNet consisting of \num{41952} synthetic 3D meshes of \num{55} categories, \eg `chair', `guitar', `airplane'.
We set the number of Transformer blocks to \num{12} with an internal dimension of \num{384}. To pre-train our point-based approach, we uniformly sample \num{8192} points from the surfaces of the objects and then resample \num{1024} points using farthest point sampling~\cite{qi2017pointnetplusplus}.
During the point cloud embedding step, we sample $n=64$ center points and $k=32$ nearest neighbors.
We train \name{} with a batch size of \num{512} for \num{800} epochs using the AdamW~\cite{loshchilov2018adamw} optimizer and a cosine learning rate decay~\cite{loshchilov2017ICLR} with a maximum learning rate of $10^{-3}$ after \num{80} epochs of linear warm-up.
For \datavec{}, we increase the batch size and learning rate to \num{2048} and $2\times10^{-3}$, respectively, as this empirically led to better results.
Following data2vec~\cite{baevski2022data2vec}, we set $\beta=2$ for the Smooth L1 loss and average the last $K=6$ blocks of the teacher.
We use minimal data augmentations during pre-training: we randomly scale the input by a factor in $[0.8, 1.2]$ and rotate it around the gravity axis.
Pre-training takes roughly \SI{18}{\hour} on a single V100 GPU.
\begin{table}
    \centering
    \caption{
        \textbf{Shape classification on ModelNet40~\cite{wu2015modelnet40}.}
        We report the overall accuracy with and without voting, as well as the mean per-class accuracy (mAcc).
    }
    \label{tab:modelnet_results}
    \small
    \setlength{\tabcolsep}{4pt}
    \renewcommand{\arraystretch}{1.2}
    \begin{tabular*}{\linewidth}{@{\extracolsep{\fill}}l *{3}{S[table-format=2.1]}@{}}
        \toprule
        \multirow{2}{*}[-0.5\dimexpr\aboverulesep+\belowrulesep+\cmidrulewidth\relax]{Method} & \multicolumn{2}{c}{Overall accuracy} & \multicolumn{1}{c}{\multirow{2}{*}[-0.5\dimexpr\aboverulesep+\belowrulesep+\cmidrulewidth\relax]{mAcc}} \\
        \cmidrule(lr){2-3}
         & \multicolumn{1}{c}{Voting} & \multicolumn{1}{c}{No voting} & \\
        \midrule
        Transf.-OcCo~\cite{yu2021pointbert} & 92.1 & {---} & {---} \\
        ParAE~\cite{eckart2021parae} & {---} & 92.9 & {---} \\
        STRL~\cite{huang2021strl} & 93.1 & {---} & {---} \\
        Point-BERT~\cite{yu2021pointbert} & 93.2 & 92.7 & {---} \\
        PointGLR~\cite{rao2020pointglr} & {---} & 93.0 & {---} \\
        OcCo~\cite{wang2021occo} & 93.0 & {---} & {---} \\
        MaskPoint~\cite{liu2022maskpoint} & 93.8 & {---} & {---} \\
        Point-MAE~\cite{pang2022pointmae} & 93.8 & 93.2 & {---} \\
        Point-M2AE~\cite{zhang2022pointm2ae} & 94.0 & 93.4 & {---} \\
        \litemidrule
        from scratch & 93.3 & 93.0 & 89.8 \\
        \datavec{} & 93.6 & 93.3 & 90.5 \\
        \textbf{\name{}} (ours) & \bfseries 94.8 & \bfseries 94.7 & \bfseries 92.0 \\
        \bottomrule
    \end{tabular*}
\end{table}

\begin{table*}[t]
    \caption{\textbf{Part segmentation results on ShapeNetPart~\cite{yi16siggraph}.}
        We report category-averaged mIoU$_C$ and instance-averaged mIoU$_I$, as well as the mean IoU for each object category.
    }
    \label{tab:ShapeNetPart}
    \centering
    \footnotesize
    \setlength{\tabcolsep}{1.25pt}
    \renewcommand{\arraystretch}{1.2}
    \begin{tabular*}{\textwidth}{@{\extracolsep{\fill}}l *{18}{S[table-format=2.1]}@{}}
        \toprule
        Method & \multicolumn{1}{c}{mIoU$_C$} & \multicolumn{1}{c}{mIoU$_I$} & \multicolumn{1}{c}{Aero} & \multicolumn{1}{c}{Bag} & \multicolumn{1}{c}{Cap} & \multicolumn{1}{c}{Car} & \multicolumn{1}{c}{Chair} & \multicolumn{1}{c}{Earph} & \multicolumn{1}{c}{Guitar} & \multicolumn{1}{c}{Knife} & \multicolumn{1}{c}{Lamp} & \multicolumn{1}{c}{Laptop} & \multicolumn{1}{c}{Motor} & \multicolumn{1}{c}{Mug} & \multicolumn{1}{c}{Pistol} & \multicolumn{1}{c}{Rocket} & \multicolumn{1}{c}{Skateb} & \multicolumn{1}{c}{Table} \\
        \midrule
            Transf.-OcCo~\cite{yu2021pointbert} & 83.4 & 85.1 & 83.3 & 85.2 & 88.3 & 79.9 & 90.7 & 74.1 & 91.9 & 87.6 & 84.7 & 95.4 & 75.5 & 94.4 & 84.1 & 63.1 & 75.7 & 80.8 \\
        Point-BERT~\cite{yu2021pointbert}    & 84.1          & 85.6          & 84.3 & 84.8 & 88.0 & 79.8 & 91.0  & 81.7  & 91.6   & 87.9  & 85.2 & 95.6   & 75.6  & 94.7 & 84.3   & 63.4   & 76.3   & 81.5  \\
            MaskPoint~\cite{liu2022maskpoint} & 84.4 & 86.0 & 84.2 & 85.6 & 88.1 & 80.3 & 91.2 & 79.5 & 91.9 & 87.8 & 86.2 & 95.3 & 76.9 & 95.0 & 85.3 & 64.4 & 76.9 & 81.8 \\
        Point-MAE~\cite{pang2022pointmae}    & 84.1          & 86.1          & 84.3 & 85.0 & 88.3 & 80.5 & 91.3  & 78.5  & 92.1   & 87.4  & 86.1 & 96.1   & 75.2  & 94.6 & 84.7   & 63.5   & 77.1   & 82.4  \\
        Point-M2AE~\cite{zhang2022pointm2ae} & \bfseries 84.9 & \bfseries 86.5 & {---}   & {---}   & {---}   & {---}   & {---}    & {---}    & {---}     & {---}    & {---}   & {---}     & {---}    & {---}   & {---}     & {---}     & {---}     & {---}    \\
        \litemidrule
        from scratch                          & 84.1          & 85.7          & 84.2 & 84.8 & 87.9 & 80.2 & 91.2  & 79.3  & 91.6   & 87.8  & 85.4 & 96.1   & 75.8  & 94.7 & 85.0   & 64.7   & 75.4   & 81.1  \\
        \datavec{}  & 84.1          & 85.9          & 84.7 & 85.8 & 87.4 & 80.7 & 91.4  & 79.3  & 91.7   & 87.9  & 85.2 & 95.8   & 76.2  & 94.9 & 83.6   & 61.6   & 78.5   & 80.7  \\
        \textbf{\name{}} (ours)               & 84.6          & 86.3          & 85.2 & 85.7 & 88.5 & 80.3 & 91.5  & 77.3  & 91.8   & 88.5  & 85.7 & 96.1   & 77.3  & 95.5 & 84.9   & 66.2   & 77.0   & 82.0  \\
        \bottomrule
    \end{tabular*}
\end{table*}

\subsection{Main Results on Downstream Tasks}
\label{sec:main_results}
In order to evaluate the effectiveness of \name{}'s self-supervised learning capabilities, we test \name{} against top-performing self-supervised methods on four different downstream tasks on well-established benchmarks.
To that end, we discard the teacher network as well as the decoder and append a task-specific head to the student network.
We then fine-tune the full network end-to-end for the specific task.
We provide detailed hyperparameters for all downstream tasks in the supplementary material.
\paragraph{Synthetic shape classification.}
\begin{figure}[t]
\centering

\resizebox{\linewidth}{!}{\input{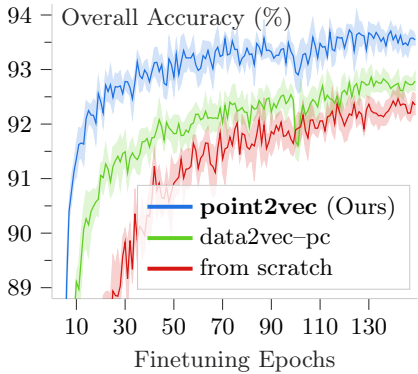}}

\caption{
\textbf{Learning curves for ModelNet40~\cite{wu2015modelnet40}.}
We show the mean (solid line) and the standard deviation (shaded background) over \num{6} independent runs of \name{}, \datavec{} as well as the model trained from scratch on ModelNet40.
\name{} consistently outperforms the baselines by a large margin.}

\label{fig:oa_learning_curve_modelnet40}
\end{figure}

\begin{table}
    \centering
    \setlength{\tabcolsep}{4pt}
    \caption{
        \textbf{Shape classification on ScanObjectNN~\cite{uy2019scanobjectnn}.}
        We report the overall accuracy over the three subsets OBJ-BG, OBJ-ONLY, and the most challenging variant PB-T50-RS.
    }
    \label{tab:scanobjectnn_results}
    \small

    \renewcommand{\arraystretch}{1.2}
    \begin{tabularx}{\linewidth}{@{}L *{3}{S[table-format=2.1]}@{}}
        \toprule
        \multirow{2}{*}[-0.5\dimexpr\aboverulesep+\belowrulesep+\cmidrulewidth\relax]{Method} & \multicolumn{3}{c}{Overall accuracy} \\
        \cmidrule(lr){2-4}
         & \multicolumn{1}{c}{OBJ-BG} & \multicolumn{1}{c}{OBJ-ONLY} & \multicolumn{1}{c}{PB-T50-RS} \\
        \midrule
        Transf.-OcCo~\cite{yu2021pointbert} & 84.9 & 85.5 & 78.8 \\
        Point-BERT~\cite{yu2021pointbert} & 87.4 & 88.1 & 83.1 \\
        MaskPoint~\cite{liu2022maskpoint} & 89.3 & 89.7 & 84.6 \\
        Point-MAE~\cite{pang2022pointmae} & 90.0 & 88.3 & 85.2 \\
        Point-M2AE~\cite{zhang2022pointm2ae} & \bfseries 91.2 & 88.8 & 86.4 \\
        \litemidrule
        from scratch & 88.1 & 88.8 & 84.3 \\
        \datavec{} & 89.7 & 88.1 & 85.5 \\
        \textbf{\name{}} (ours) & \bfseries 91.2 & \bfseries 90.4 & \bfseries 87.5 \\
        \bottomrule
    \end{tabularx}

\end{table}

After pre-training on ShapeNet, we fine-tune our model for shape classification on ModelNet40~\cite{wu2015modelnet40}, which consists of \num{12311} synthetic 3D models of \num{40} semantic categories.
We obtain the semantic class label by passing the concatenated mean- and max-pooled output of the Transformer encoder into a \num{3}-layer MLP and fine-tune the whole network end-to-end.
We use minimal data augmentations consisting only of resampling \num{1024} points with farthest point sampling, applying random anisotropic scaling of up to $40\%$, centering at the origin, and rescaling to the unit sphere.
Other commonly used augmentations did not improve performance; \eg random rotations around the axis of gravity and random translations are detrimental as ModelNet40 instances are canonically oriented.
During the point cloud embedding step, we sample $n=64$ center points and $k=32$ nearest neighbors.
In~\cref{tab:modelnet_results}, we report state-of-the-art results for shape classification on ModelNet40 among self-supervised methods by a margin of \num{1.3} percentage points without voting~\cite{zhang2022pointm2ae, pang2022pointmae, yu2021pointbert}.
Interestingly, pre-training with \datavec{} results in only marginal improvements (\num{0.3} percentage points without voting) over the same model trained from scratch on ModelNet40.
Unlike \datavec{}, we observe that \name{} unleashes the full potential of data2vec-like pre-training on ModelNet40 by achieving substantial performance gains of \num{1.7} percentage points over the baseline trained from scratch.
In~\cref{fig:oa_learning_curve_modelnet40},
we plot the accuracy per training epoch of \name{}, \datavec{}, as well as our baseline trained from scratch on ModelNet40.
We observe that \name{} outperforms our strong baselines by a consistent margin throughout the entire training.
\name{} effectively learns strong feature representations on ShapeNet, resulting in a significantly accelerated adaptation to the fine-tuning task (\cref{fig:oa_learning_curve_modelnet40}).\paragraph{Real-world shape classification.}
\begin{table}[t]
    \caption{
        \textbf{Few-shot classification on ModelNet40~\cite{wu2015modelnet40}.}
        We report the mean and standard deviation over \num{10} runs.
    }
    \label{tab:modelnet_fewshot_results}
    \centering
    \footnotesize
    \setlength{\tabcolsep}{1.5pt}
    \renewcommand{\arraystretch}{1.2}
    \begin{tabularx}{\linewidth}{@{}L *{4}{S[table-format=2.1(2),uncertainty-mode=separate,table-align-uncertainty=false]}@{}}
        \toprule
        \multirow{2}{*}[-0.5\dimexpr\aboverulesep+\belowrulesep+\cmidrulewidth\relax]{Method} & \multicolumn{2}{c}{5-way} & \multicolumn{2}{c}{10-way} \\
        \cmidrule(lr){2-3} \cmidrule(lr){4-5}
         & \multicolumn{1}{c}{10-shot} & \multicolumn{1}{c}{20-shot} & \multicolumn{1}{c}{10-shot} & \multicolumn{1}{c}{20-shot} \\
        \midrule
        OcCo~\cite{wang2021occo} & 91.9(36) & 93.9(31) & 86.4(54) & 91.3(46) \\
        Transf.-OcCo~\cite{yu2021pointbert} & 94.0(36) & 95.9(23) & 89.4(51) & 92.4(46) \\
        Point-BERT~\cite{yu2021pointbert}       & 94.6(31)             & 96.3(27)               & 91.0(54)               & 92.7(51)               \\
        MaskPoint~\cite{liu2022maskpoint} & 95.0(37) & 97.2(17) & 91.4(40) & 93.4(35) \\
        Point-MAE~\cite{pang2022pointmae}                     & 96.3(25)           & 97.8(18) & 92.6(41) & 95.0(30) \\
        Point-M2AE~\cite{zhang2022pointm2ae}                     & 96.8(18)           & 98.3(14) & 92.3(45) & 95.0(30) \\
        \litemidrule
        from scratch                    & 93.8(32)           & 97.1(19)             & 90.1(46)             & 93.6(39)             \\
        \datavec{}  & 96.2(26)           & 97.8(22)             & 92.6(49)             & 95.0(32) \\
        \textbf{\name{}} (ours) & \bfseries 97.0(28)  & \bfseries 98.7(12)    & \bfseries 93.9(41)    & \bfseries 95.8(31)    \\
        \bottomrule
    \end{tabularx}

\end{table}

Next, we fine-tune \name{} on ScanObjectNN~\cite{uy2019scanobjectnn}, which contains \num{2902} real-world object scans of \num{15} semantic classes.
In contrast to shape classification on ModelNet40, we do not resample points but use all \num{2048} points and sample $n=128$ center points for the point cloud embedding step. We found more aggressive scaling to be detrimental and use random anisotropic scaling of up to $10\%$. \Cref{tab:scanobjectnn_results} shows that, although pre-trained on synthetic data, \name{} generalizes well to cluttered real-world data and achieves state-of-the-art performance among self-supervised methods by a margin of \num{1.1} percentage points on PB-T50-RS, the most difficult variant of the dataset.
We observe that pre-training \name{} on ShapeNet plays a crucial role in its strong performance.
Compared to the baseline trained from scratch on ScanObjectNN, pre-training with \name{} achieves a performance gain of \num{3.2} percentage points.
We again report improvements of \name{} over \datavec{} of up to \num{2.3} percentage points.

\paragraph{Few-shot classification.}
Following the few-shot evaluation protocol of \citet{sharma2020ssl}, we test the few-shot capabilities of \name{} in an $m$-way, $n$-shot setting.
To this end, we randomly sample $m$ classes and select $n$ instances for training at random for each of these classes.
For testing, we randomly pick \num{20} unseen instances from each of the $m$ support classes.
Following prior work~\cite{pang2022pointmae}, we report the mean and standard deviation over \num{10} independent runs.
In \cref{tab:modelnet_fewshot_results}, we report state-of-the-art results, with an improvement of \num{1.3} percentage points in the most difficult \num{10}-way \num{10}-shot setting.
\name{} clearly outperforms the \datavec{} baseline in all settings.
We conclude that \name{} learns rich feature representations which are also well suited for transfer learning in a low-data regime.

\paragraph{Part segmentation.}
Finally, we address the task of part segmentation, which assigns a semantic part label to each point in a 3D point cloud of a single object.
For this purpose, we employ a simple segmentation head that is similar to the segmentation head in Point-MAE.
First, we average the outputs of the 4th, 8th, and 12th Transformer blocks to incorporate features from multiple levels of abstraction.
We then concatenate the mean- and max-pooled features of the $n$ averaged token outputs, along with the one-hot encoded class label of the object, to obtain a global feature vector.
At the same time, we upsample the $n$ averaged outputs from the corresponding center points to all points using a PointNet++~\cite{qi2017pointnetplusplus} feature propagation layer, which uses inverse distance weighting and a shared MLP to produce a local feature vector for each point.
Finally, we concatenate the global feature vector with each local feature vector, and a shared MLP predicts a part label for each point.
In~\cref{tab:ShapeNetPart}, we report competitive results on ShapeNetPart~\cite{yi16siggraph}, which consists of \num{16881} 3D models of \num{16} semantic categories.
Apart from Point-M2AE, \name{} outperforms all other self-supervised methods.
We hypothesize that Point-M2AE's multi-scale U-Net-like architecture~\cite{ronneberger2015unet} enables it to learn more expressive spatially localized features, resulting in slightly better scores ($+0.2$\,mIoU$_I$).
Since \name{} relies on a standard single-scale Transformer backbone, we see multi-scale Transformers for 3D point clouds as an interesting orthogonal improvement, similar to the advances in 2D vision~\cite{zhang2021longformer,fan2021msvit,li2022msvit2,chen2021crossvit} extending vision Transformers~\cite{dosovitskiy2020vit} with multi-scale capabilities.
\subsection{Analysis}
\label{sec:analysis}
\begin{table}

    \centering
    \setlength{\tabcolsep}{4pt}
    \caption{
        \textbf{Ablation.}
        We find that a deferred shallow decoder~(D) (\cref{fig:model},~\colorsquare{m_red}) predicting the teacher's representations for masked patches shows consistent improvements but we identify that concealing positional information (No \protect\maskembedding{}) from the student is key.
        }
    \label{tab:architecture_ablation}

    \small
    \renewcommand{\arraystretch}{1.2}
    \begin{tabular*}{\linewidth}{@{\extracolsep{\fill}}l cc *{3}{S[table-format=2.1]}@{}}
        \toprule
         &  &  & \multicolumn{3}{c}{Overall accuracy} \\
        \cmidrule(lr){4-6}
        Method & No \protect\maskembedding{} & D & \multicolumn{2}{c}{ModelNet40} & \multicolumn{1}{c}{ScanObjNN} \\
        \cmidrule(lr){4-5} \cmidrule(lr){6-6}
         &  &  & \multicolumn{1}{c}{Voting} & \multicolumn{1}{c}{No voting} & \multicolumn{1}{c}{\small PB-T50-RS} \\
        \midrule
                    \datavec{} &\xmark  & \xmark           & 93.6                          & 93.3             & 85.5             \\
                     &\xmark & \cmark & 94.0              & 93.6 & 86.8             \\
                \textbf{\name{}} (ours) &\cmark & \cmark & \bfseries 94.8                 & \bfseries 94.7    & \bfseries 87.5    \\
        \bottomrule
    \end{tabular*}

\end{table}

\paragraph{Leakage of positional information.}
\begin{figure}[t]
\centering
\subcaptionbox{\scriptsize Disclosed positions: \datavec{}\label{fig:early_leakage_disclosed}}{\includegraphics[width=0.50\linewidth,trim={0cm 0.8cm 0cm, 1.4cm},clip]{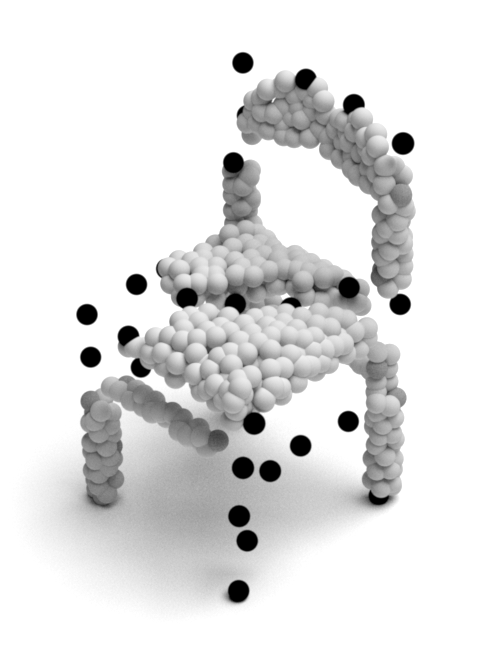}}\hfill
\subcaptionbox{\scriptsize Concealed positions: \textbf{\name{}} (ours)\label{fig:early_leakage_concealed}}{\includegraphics[width=0.50\linewidth,trim={0cm 0.8cm 0cm, 1.4cm},clip]{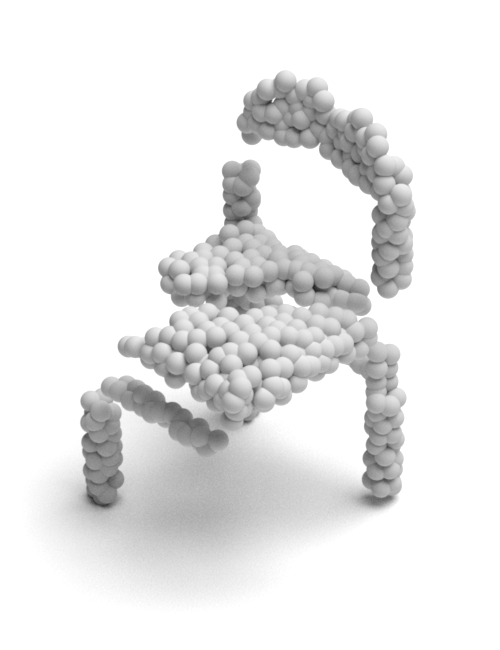}}\caption{\textbf{Leakage of positional information.}
The center points~\colordot{black} of masked point patches are associated with the masked embeddings (\cref{fig:model},~\protect\maskembedding{}).
(a) \datavec{} discloses the positions of masked patches to the student, revealing the chair's overall shape.
(b) \name{} excludes masked embeddings from the student and therefore conceals the positions of the masked patches.
}
\label{fig:early_leakage}
\end{figure}

The main limitation of \datavec{} is that it directly feeds masked embeddings, along with their positional information, to the student network, which undermines the effectiveness of masking.
To visualize this problem, we show a representative example in \cref{fig:early_leakage_disclosed}.
Revealing the positions of masked patches of the chair inadvertently weakens the learning objective because it allows the student to rely on the positional information instead of truly learning to predict the teacher's representations of the corresponding masked-out patches.
To mitigate this issue, \name{} excludes masked embeddings from the student and only subsequently feeds them to the decoder.
As a result, several sections of the chair in \cref{fig:early_leakage_concealed} are effectively concealed from the student network, leading to a more resilient learning framework.
In \cref{tab:architecture_ablation}, we report that \name{} outperforms our baseline \datavec{} by a margin of up to \num{2.0} percentage points.
In particular, we observe that the decoder itself provides consistent improvements, but the key contribution of \name{} is to conceal positional information from the student, \ie shifting mask tokens from the encoder's input to the decoder (No~\protect\maskembedding{}).
Complementary to our findings, \citet{he2022mae} show that moving masked embeddings to a deferred shallow decoder reduces memory requirements and training time significantly.
Our findings align with those of \citet{pang2022pointmae}, who found similar benefits for masked autoencoders on point clouds.

\begin{figure*}[t!]
    \includegraphics[width=1.0\linewidth,trim={0cm 0.65cm 0cm 0.8cm},clip]{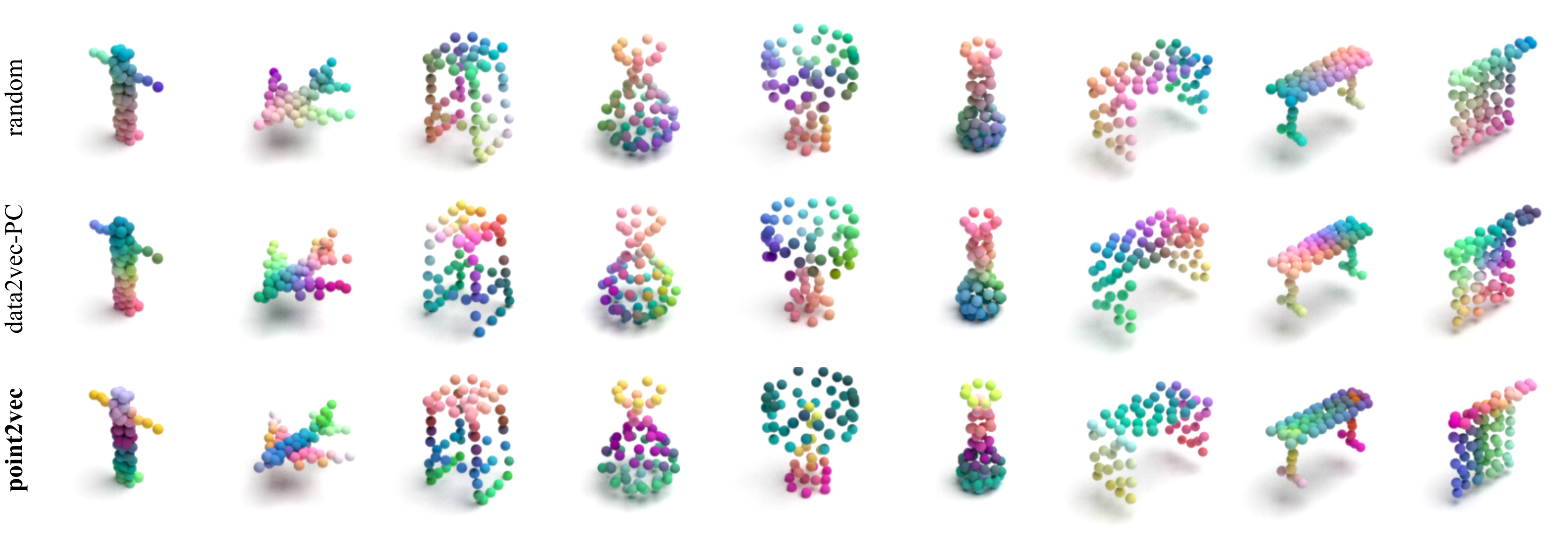}

    \caption{
        \textbf{Visualization of learned representations.}
        We use PCA to project the learned representations into RGB space.
        Both a random initialization and \datavec{} pre-training show a fairly strong positional bias, whereas \name{} exhibits stronger semantic grouping without being trained on downstream dense prediction tasks.
    }
    \label{fig:quali_representations}

\end{figure*}

\paragraph{Masking strategy.}
\begin{table}
    \centering
    \caption{
        \textbf{Masking strategy.}
        We explore two variants for masking the input of the student.
        For (a) random masking, we uniformly mask out a given ratio of all embeddings.
        For (b) block masking, we mask out a random embedding and its nearest neighbors.
        We report the overall accuracy on ModelNet40 and ScanObjectNN.}
    \label{tab:ablation_masking_ratio}

    \subcaptionbox{$65\%$ random masking}[0.48\linewidth]{\includegraphics[width=\linewidth]{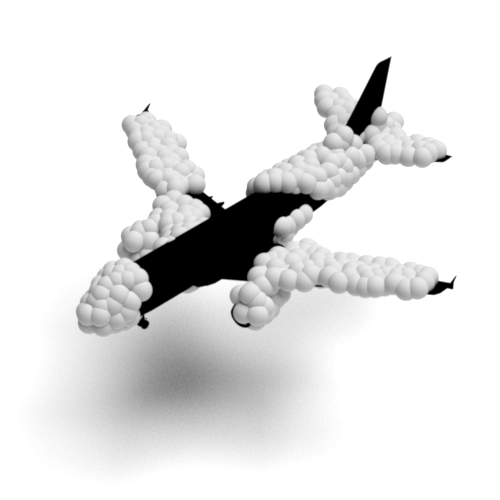}}\hfill
    \subcaptionbox{$65\%$ block masking}[0.48\linewidth]{\includegraphics[width=\linewidth]{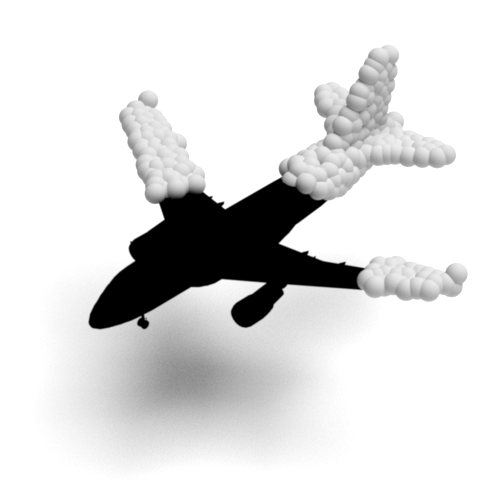}}

    \setlength{\tabcolsep}{4pt}
    \small
    \renewcommand{\arraystretch}{1.2}
    \begin{tabular*}{\linewidth}{@{\extracolsep{\fill}}l S[table-format=2.0] *{3}{S[table-format=2.1]}@{}}
        \toprule
         &  & \multicolumn{3}{c}{Overall accuracy} \\
        \cmidrule(lr){3-5}
        Strategy & \multicolumn{1}{c}{Ratio (\%)} & \multicolumn{2}{c}{ModelNet40} & \multicolumn{1}{c}{ScanObjNN} \\
        \cmidrule(lr){3-4} \cmidrule(lr){5-5}
         &  & \multicolumn{1}{c}{Voting} & \multicolumn{1}{c}{No voting} & \multicolumn{1}{c}{\small PB-T50-RS} \\
        \midrule
        random                      & 45    & 94.5                          & 94.3          & 86.8          \\ random & 65    & \bfseries 94.8                 & \bfseries 94.7 & \bfseries 87.5 \\ random                      & 85    & 94.5                          & 93.8           & 86.7          \\ \litemidrule
        block                       & 25    & 93.9                          & 93.7          & 86.3          \\ block  & 45    & 94.5                & 93.8       & 87.4 \\ block                       & 65    & 94.0                          & 93.9 & 86.1          \\ \bottomrule
    \end{tabular*}

\end{table}

The masking strategy defines which of the student's input embeddings are masked (\cref{fig:model},~\protect\maskembedding{}).
In this study, we investigate two variants of masking strategies with different masking ratios: \emph{random} masking and \emph{block} masking.
For random masking, we mask out a specified ratio of embeddings for the student.
In contrast, block masking masks out a random embedding and its nearest neighbors such that the specified masking ratio is achieved.
This strategy focuses on masking out spatially contiguous regions of the point cloud, whereas random masking is independent of position.
Our findings, summarized in \cref{tab:ablation_masking_ratio}, reveal that random masking with a $65\%$ masking ratio performs best for both ModelNet40 and ScanObjectNN, while block masking lags behind.
We attribute this to the high level of ambiguity that arises when masking a spatially contiguous region, resulting in several potential point clouds that could have given rise to the masked input.
While we seek a challenging pretext task to learn rich representations, ambiguity should not be the primary source of difficulty.

Recall that our masking strategy is applied to patch embeddings rather than individual points.
Consequently, points may belong to both masked and unmasked patches.
While certain masked patches may be easy to predict, our masking ratio of $65\%$ ensures that there are still plenty of regions entirely masked (\cref{fig:early_leakage,tab:ablation_masking_ratio}).
As a result, we conclude that the masking ratio is a sensitive hyperparameter that requires some careful tuning to strike the right balance of difficulty for the pretext task. \paragraph{Visualization of representations learned by \name{}.}
In \cref{fig:quali_representations}, we show qualitative examples of representations of ModelNet40 instances after pre-training on ShapeNet.
Both a random initialization and \datavec{} pre-training show a strong positional bias, whereas \name{} exhibits stronger semantic grouping without being trained on downstream dense prediction tasks.
Unlike \datavec{}, \name{} conceals positional information from the student, forcing it to learn more about the semantics of the data, resulting in more semantically meaningful representations.
\section{Conclusion}
In this work, we have extended data2vec to the point cloud domain.
Through an in-depth analysis, we have discovered that the disclosure of positional information to the student network limits data2vec's ability to learn strong representations on point clouds.
To overcome this limitation, we have introduced \name{}, a self-supervised representation learning approach which unleashes the full potential of data2vec-like pre-training on point clouds.
\name{} achieves remarkable results on various downstream tasks, surpassing other self-supervised learning approaches in few-shot learning as well as shape classification on well-established benchmarks.
Future work might include extending \name{} to scene-level representation learning. 

{\paragraph{Acknowledgments.}
This work has been funded, in part, by the ERC CoG grant DeeViSe (\texttt{ERC-2017-CoG-773161}) and by the BMBF project 6GEM (\texttt{16KISK036K}).
We gratefully acknowledge computing resources granted by RWTH Aachen University (\texttt{thes1313}, \texttt{supp0003}) and thank István Sárándi for helpful discussions and feedback.
}

{\small
\bibliographystyle{ieeenat_fullname}
\bibliography{main}
}

\clearpage
\appendix
\maketitlesupplementary
\noindent
We provide additional ablation studies, architecture and training details, and qualitative part-segmentation results.
Our code and model are publicly available for research purposes.

\section{Further Ablation Studies}
\paragraph{Pretext task.}
In the main paper, we have only explored self-supervised pre-training on the ShapeNet dataset~\cite{chang2015shapenet}.
However, ShapeNet also contains class labels which could instead be used for fully supervised classification-based pre-training.
As can be seen in \cref{tab:pretext_tasks}, this performs substantially worse than \name{} and offers no improvement over training from scratch.
We can also pre-train using \name{} directly on ModelNet40~\cite{wu2015modelnet40}, which constitutes roughly a quarter of ShapeNet's size.
Still, we see improved downstream performance, indicating that the \name{} pretext task is meaningful for pre-training.

\paragraph{EMA rate warm-up.}
During pre-training, we linearly warm up the EMA rate $\tau$ over the first $\tau_n$ epochs from its initial value $\tau_0$ to its final value $\tau_e$~\cite{baevski2022data2vec}.
This approach is based on the idea that we should update the teacher network more rapidly at the start of training since the feature representations are not yet well established.
In~\cref{tab:ema_tau_ablation}, we show overall accuracy scores on ModelNet40 and the PB-T50-RS variant of ScanObjectNN~\cite{uy2019scanobjectnn} using various values for $\tau_n$.
Our findings suggest that $\tau_n$ is a crucial hyperparameter for achieving effective pre-training with \name{}.
In~\cref{tab:hyperparameters}, we provide the EMA rates employed by our baseline \datavec{} and \name{}, respectively.
\begin{table}
    \centering
    \setlength{\tabcolsep}{4pt}
    \caption{
        \textbf{Pretext tasks.}
        Pre-training with a classification objective on ShapeNet offers no gains over training from scratch on ModelNet40 and yields worse performance on the PB-T50-RS variant of ScanObjectNN.
        However, \name{} already brings performance gains when pre-trained with the much smaller ModelNet40 dataset and significant improvements when pre-trained with the large ShapeNet dataset.
    }
    \label{tab:pretext_tasks}
    \small
    \renewcommand{\arraystretch}{1.2}
    \begin{tabular*}{\linewidth}{@{\extracolsep{\fill}}l *{3}{S[table-format=2.1]}@{}}
        \toprule
         & \multicolumn{3}{c}{Overall accuracy} \\
        \cmidrule(lr){2-4}
        Pretext task & \multicolumn{2}{c}{ModelNet40} & \multicolumn{1}{c}{ScanObjNN} \\
        \cmidrule(lr){2-3} \cmidrule(lr){4-4}
         & \multicolumn{1}{c}{Voting} & \multicolumn{1}{c}{No voting} & \multicolumn{1}{c}{\small PB-T50-RS} \\
        \midrule
            none / from scratch & 93.3 & 93.0 & 84.3 \\
            classification (ShapeNet) & 93.2                                    & 93.0 & 82.9 \\
            \litemidrule
            \name{} (ModelNet40) & 93.9 & 93.6 & 84.4 \\
        \name{} (ShapeNet) & \bfseries 94.8 & \bfseries 94.7 & \bfseries 87.5 \\
            \bottomrule
    \end{tabular*}
\end{table}

\begin{table}
    \centering
    \caption{
        \textbf{EMA rate warm-up.}
        We linearly warm up the EMA rate during the first $\tau_n$ epochs.
    }
    \label{tab:ema_tau_ablation}
    \small
    \renewcommand{\arraystretch}{1.2}
    \begin{tabular*}{\linewidth}{@{\extracolsep{\fill}}S[table-format=3.0] *{3}{S[table-format=2.1]}@{}}
        \toprule
         & \multicolumn{3}{c}{Overall accuracy} \\
        \cmidrule(lr){2-4}
        \multicolumn{1}{c}{$\tau_n$} & \multicolumn{2}{c}{ModelNet40} & \multicolumn{1}{c}{ScanObjNN} \\
        \cmidrule(lr){2-3} \cmidrule(lr){4-4}
         & \multicolumn{1}{c}{Voting} & \multicolumn{1}{c}{No voting} & \multicolumn{1}{c}{\small PB-T50-RS} \\
        \midrule
        80       & 94.5                                    & 94.1               & 86.7 \\
            160      & 94.6                                    & 94.2               & 87.4 \\
            200      & \bfseries 94.8                           & \bfseries 94.7      & \bfseries 87.5 \\
            300      & 94.1                                    & 94.0               & 87.3 \\
            400      & 94.0                                    & 94.0               & 87.3 \\
            \bottomrule
    \end{tabular*}
\end{table}

\begin{figure*}[t!]

\includegraphics[width=1.0\linewidth,trim={0cm 0.85cm 0cm 0.85cm},clip]{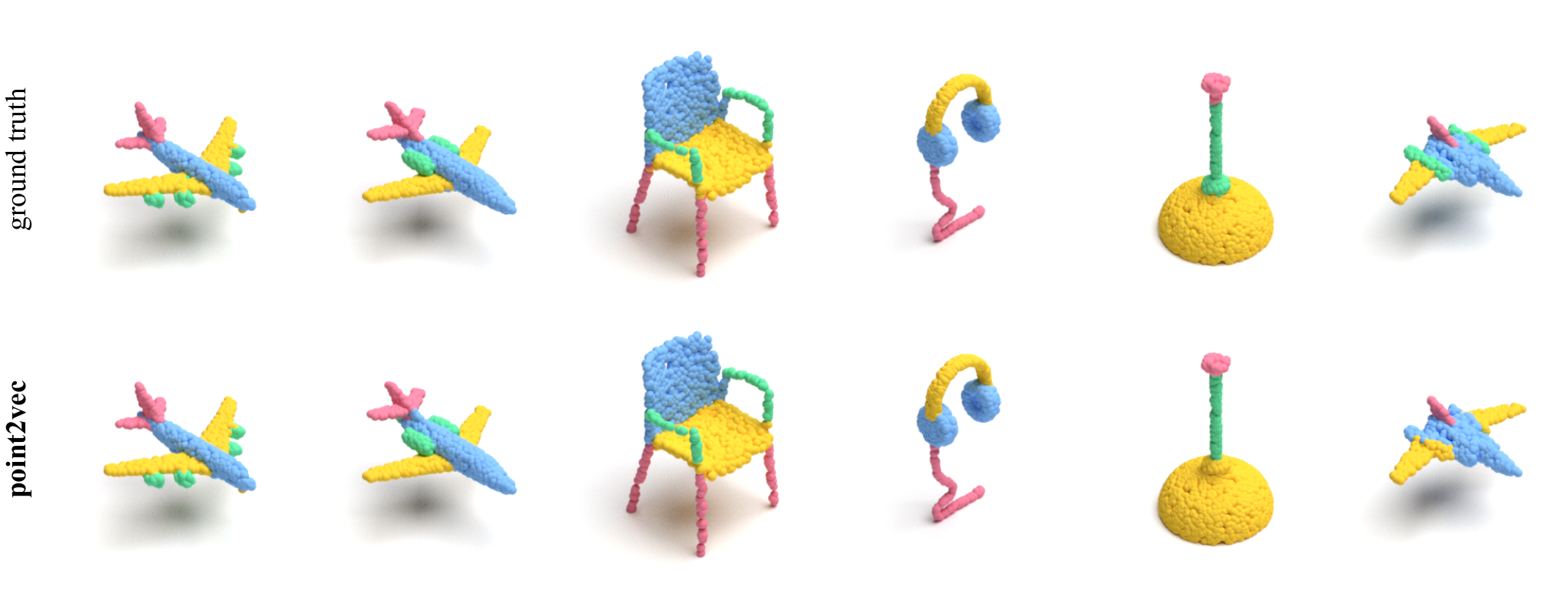}
\caption{
\textbf{Qualitative results on ShapeNetPart~\cite{yi16siggraph}.}
\name{} produces well-localized boundaries between parts with minimal semantic errors.
In most cases, the differences between the results of \name{} and the ground truth are imperceptible to the human eye.
However, the last example shows a failure case where the jet engine is not properly segmented.
}
\label{fig:shapenetpart_qualitative}
\end{figure*}

\paragraph{Target layer aggregation.}
During training of \datavec{}, as well as \name{}, we need to specify which layers of the teacher to use for the target.
The target is constructed by averaging the last $K$ layers of the teacher; \citet{baevski2022data2vec} use $K=6$ for their vision models.
We ablate this hyperparameter and report results in \cref{tab:number_of_target_layers}.
Although all tested values lead to usable results, $K=6$ gives the best overall performance.
\begin{table}[t]
    \centering
    \caption{
        \textbf{Target layer aggregation.}
        We construct training targets by averaging the outputs of the last $K$ Transformer blocks of the teacher.
        We observe that $K=6$ is optimal for ModelNet40 and close to optimal for the PB-T50-RS variant of ScanObjectNN.
    }
    \label{tab:number_of_target_layers}
    \small
    \renewcommand{\arraystretch}{1.2}
    \begin{tabular*}{\linewidth}{@{\extracolsep{\fill}}S[table-format=2.0] *{3}{S[table-format=2.1]}@{}}
        \toprule
         & \multicolumn{3}{c}{Overall accuracy} \\
        \cmidrule(lr){2-4}
        \multicolumn{1}{c}{$K$} & \multicolumn{2}{c}{ModelNet40} & \multicolumn{1}{c}{ScanObjNN} \\
        \cmidrule(lr){2-3} \cmidrule(lr){4-4}
         & \multicolumn{1}{c}{Voting} & \multicolumn{1}{c}{No voting} & \multicolumn{1}{c}{\small PB-T50-RS} \\
        \midrule
        1      & 94.4                                    & 94.1               & 87.0 \\
            3      & 94.7                                    & 94.3               & 87.1 \\
            6      & \bfseries 94.8                           & \bfseries 94.7      & 87.5 \\
            9      & 94.3                                    & 94.0               & \bfseries 87.6 \\
            12     & 94.5                                    & 94.3               & 87.3 \\
            \bottomrule
    \end{tabular*}
\end{table}

\begin{table}[t]
        \centering
    \caption{
        \textbf{Hyperparameters for part segmentation.}
            We use the same hyperparameters when fine-tuning \name{} and \datavec{} on ShapeNetPart~\cite{yi16siggraph}.
    }
    \label{tab:hyperparameters_part_segmentation}
    \small
    \setlength{\tabcolsep}{4pt}
    \renewcommand{\arraystretch}{1.2}
    \begin{tabularx}{\linewidth}{@{}L l@{}}
        \toprule
        Parameter & Value \\
        \midrule
        Epochs & \num{300} \\
        Batch size & \num{16}                  \\
        Optimizer & AdamW               \\
        Learning rate & $2 \times 10^{-4}$  \\
        Weight decay & \num{0.05}                \\
        LR schedule & cosine              \\
        LR warm-up & \num{10} epochs           \\
        \litemidrule
        Points & \num{2048} \\
        $n$ (center points) & \num{128} \\
        $k$ ($k$-NN grouping) & \num{32} \\
        mini-PointNet MLP 1 & \num{128}, \num{256}                  \\
        mini-PointNet MLP 2 & \num{512}, \num{384}                  \\
        \litemidrule
        Encoder layers & \num{12}                  \\
        Encoder dimension & \num{384}                 \\
        Encoder heads & \num{6}                 \\
        Encoder drop path & $0\%, \ldots, 20\%$                 \\
        \litemidrule
        Feature propagation & See \cref{sec:main_results} \\
        Head dimensions & \num{512}, \num{256}, \#classes \\
        Head dropout & $50\%$ \\
        \bottomrule
    \end{tabularx}

\end{table}
\begin{figure}[t]

\definecolor{baselinecurve}{HTML}{D55E00}
\definecolor{ourcurve}{HTML}{2072C9}
\pgfplotsset{efficiency axis/.style={
    width=\linewidth,
    height=4cm,
    font=\footnotesize,
    axis x line=bottom,
    axis y line=left,
    x axis line style={-},
    y axis line style={-},
    axis line style={black!45},
    tick align=outside,
    tick style={black!45},
    xticklabel style={inner xsep=0pt},
    yticklabel style={inner xsep=0pt},
    ymajorgrids=true,
    grid style={black!10},
    xlabel={Pre-training data (\%)},
    ylabel={Overall accuracy (\%)},
    xmin=0, xmax=100,
    xtick={0,25,50,75,100}
}}

\centering
{\footnotesize
\tikz[baseline=-0.5ex]\draw[ourcurve,line width=0.9pt] (0,0) -- (1.5em,0) node[midway,circle,fill,inner sep=1.5pt] {};
\textbf{\name{}} (ours)
\quad
\tikz[baseline=-0.5ex]\draw[baselinecurve,line width=0.7pt] (0,0) -- (1.5em,0) node[midway,circle,fill,inner sep=1.5pt] {};
Point-MAE\par}

\begin{minipage}{.49\linewidth}
\subcaptionbox{ModelNet40~\cite{wu2015modelnet40}}{\resizebox{\linewidth}{!}{\begin{tikzpicture}
\begin{axis}[
  efficiency axis,
  ytick={93,94,95},
  ymin=92.1, ymax=95.0
]
\addplot [line width=0.9pt, ourcurve, mark=*, mark size=1.5pt]
table {0 93.3
25 93.9
50 94.0
75 94.4
100 94.8
};
\addplot [line width=0.7pt, baselinecurve, mark=*, mark size=1.5pt]
table {0 93.0
25 93.2
50 93.8
75 93.4
100 93.8
};
\end{axis}
\end{tikzpicture}}}\end{minipage}\hfill
\begin{minipage}{.49\linewidth}
\subcaptionbox{ScanObjectNN~\cite{uy2019scanobjectnn}}{\resizebox{\linewidth}{!}{\begin{tikzpicture}
\begin{axis}[
  efficiency axis,
  ytick={82,84,86,88},
  ymin=81.1, ymax=88.0
]
\addplot [line width=0.9pt, ourcurve, mark=*, mark size=1.5pt]
table {0 84.3
25 85.7
50 86.5
75 86.7
100 87.5
};
\addplot [line width=0.7pt, baselinecurve, mark=*, mark size=1.5pt]
table {0 81.6
25 83.7
50 84.3
75 84.7
100 85.2
};
\end{axis}
\end{tikzpicture}}}\end{minipage}

\caption{
\textbf{Pre-training data efficiency.}
Irrespective of the quantity of pre-training data used from ShapeNet, \name{} consistently achieves better results than Point-MAE~\cite{pang2022pointmae} on ModelNet40 (with voting) and the PB-T50-RS variant of ScanObjectNN.
}

\label{fig:pretraining_data_efficiency}

\end{figure}

\paragraph{Pre-training data efficiency.}
We evaluate the efficiency of self-supervised pre-training with \name{}.
To this end, we partition the ShapeNet training dataset into subsets containing $25\%$, $50\%$, $75\%$, and $100\%$ of the data.
We then fine-tune our model for shape classification on ModelNet40 and ScanObjectNN.
As shown in~\cref{fig:pretraining_data_efficiency}, \name{} achieves consistent improvements on both datasets.
Notably, pre-training \name{} with only $25\%$ of the training data yields superior results compared to Point-MAE~\cite{pang2022pointmae} pre-trained with $100\%$ of the training data.

\begin{table*}[t]
    \begin{minipage}[t]{\columnwidth}
            \centering
    \caption{
        \textbf{Hyperparameters for \datavec{} and \name{}.}
        \datavec{} denotes our adaptation of data2vec to the point cloud modality.
        We report the best-performing hyperparameters for both \datavec{} and \name{}.
        LN: layer normalization. AVG: average pooling over layers.
    }
    \label{tab:hyperparameters}
    \small
    \setlength{\tabcolsep}{4pt}
    \renewcommand{\arraystretch}{1.2}
    \begin{tabularx}{\linewidth}{@{}L ll@{}}
        \toprule
        Parameter & \datavec{} & \name{} (ours) \\
        \midrule
        Epochs & \num{800} & \num{800}                                \\
        Optimizer & AdamW & AdamW                                     \\
        Learning rate & $2 \times 10^{-3}$ & $1 \times 10^{-3}$                                \\
        Weight decay & \num{0.05} & \num{0.05}                                      \\
        LR schedule & cosine & cosine                                    \\
        LR warm-up & \num{80} epochs & \num{80} epochs                       \\
        Batch size & \num{2048} & \num{512}                              \\
        \litemidrule
        Encoder layers & \num{12} & \num{12}                                        \\
        Encoder dimension & \num{384} & \num{384}                                       \\
        Decoder layers & --- & \num{4}                                         \\
        Masking strategy & random & random                                    \\
        Masking ratio & $65\%$ & $65\%$ \\
        \litemidrule
        $\tau_0$ (EMA start) & \num{0.9998} & \num{0.9998}                              \\
        $\tau_e$ (EMA end) & \num{0.99999} & \num{0.99999}     \\
        $\tau_n$ (EMA warm-up) & \num{200} epochs & \num{200} epochs                       \\
        $K$ (layers to average) & \num{6} & \num{6}                                         \\
        Target normalization & LN$\rightarrow$AVG$\rightarrow$LN & LN$\rightarrow$AVG$\rightarrow$LN         \\
        \bottomrule
    \end{tabularx}

    \end{minipage}\hfill
    \begin{minipage}[t]{\columnwidth}
            \centering
    \caption{
        \textbf{Hyperparameters for classification.}
        Shared settings for \name{} and \datavec{} on ModelNet40~\cite{wu2015modelnet40} and ScanObjectNN~\cite{uy2019scanobjectnn}. Paired values list ModelNet40 first and ScanObjectNN second.
        Training from scratch uses a learning rate of $1 \times 10^{-3}$ and no encoder freezing.
    }
    \label{tab:hyperparameters_classification}
    \small
    \setlength{\tabcolsep}{4pt}
    \renewcommand{\arraystretch}{1.2}
    \begin{tabularx}{\linewidth}{@{}L l@{}}
        \toprule
        Parameter & Value \\
        \midrule
        Epochs & \num{150} \\
        Batch size & \num{32}                  \\
        Optimizer & AdamW               \\
        Learning rate & $3 \times 10^{-4}$  \\
        Weight decay & \num{0.05}                \\
        LR schedule & cosine              \\
        LR warm-up & \num{10} epochs           \\
        \litemidrule
        Points & \num{1024} / \num{2048} \\
        $n$ (center points) & \num{64} / \num{128} \\
        $k$ ($k$-NN grouping) & \num{32} \\
        mini-PointNet MLP 1 & \num{128}, \num{256}                  \\
        mini-PointNet MLP 2 & \num{512}, \num{384}                  \\
        \litemidrule
        Encoder layers & \num{12}                  \\
        Encoder dimension & \num{384}                 \\
        Encoder heads & \num{6}                 \\
        Encoder drop path & $0\%,\ldots,20\%$                 \\
        Encoder frozen & \num{100} epochs \\
        \litemidrule
        Feature aggregation & mean + max pooling \\
        Head dimensions & \num{256}, \num{256}, \#classes \\
        Head dropout & $50\%$ \\
        Label smoothing & \num{0.2} \\
        \bottomrule
    \end{tabularx}

    \end{minipage}
\end{table*}

\section{Architecture Details}

In \cref{tab:hyperparameters}, we provide detailed hyperparameters for pre-training \datavec{} and \name{} on ShapeNet~\cite{chang2015shapenet}.
We also report the hyperparameters for fine-tuning \name{} on shape classification (\cref{tab:hyperparameters_classification}) and part segmentation (\cref{tab:hyperparameters_part_segmentation}).
In \cref{lst:point2vec}, we provide the PyTorch-inspired pseudocode for \name{} pre-training.
The decoder outputs are mapped to the target space by a learned linear projection.

\section{Qualitative Results for Part Segmentation}
\Cref{fig:shapenetpart_qualitative} shows qualitative results on ShapeNetPart~\cite{yi16siggraph}: \name{} accurately localizes part boundaries, and most predictions visually match the ground truth.

\begin{figure*}[t]
\begin{pseudocode}[\label{lst:point2vec}]{PyTorch-inspired pre-training for \name{}.}
# N: batch size (512)
# S: number of groups/embeddings (64)
# V: number of visible groups
# E: embedding feature dimension (384)

for point_cloud in data_loader:
    # point cloud embedding
    center_points = fps(point_cloud)  # (N, S, 3)
    point_patches = knn(point_cloud, center_points)  # (N, S, 32, 3)
    point_patches = point_patches - center_points.unsqueeze(dim=2)
    patch_embeddings = mini_pointnet(point_patches)  # (N, S, E) (!\cref{fig:model}!, !\colorsquare{m_pointnet}!)
    pos_embeddings = pos_encoder(center_points)  # (N, S, E)

    # masking
    mask_embeddings = mask_embedding.expand(N, S, E)  # (N, S, E)
    mask = generate_mask(center_points)  # (N, S) boolean

    # targets
    with torch.no_grad():
        teacher_states = teacher(patch_embeddings, pos_embeddings)  # (12, N, S, E) (!\cref{fig:model}!, !\colorsquare{m_green}!)
        target_layers = [F.layer_norm(x, (E,)) for x in teacher_states[6:]]  # [(N, S, E)]
        targets = torch.stack(target_layers).mean(dim=0)  # (N, S, E)
        targets = F.layer_norm(targets, (E,))  # (N, S, E)

    # predictions
    last_student_state = student(  # (N, V, E) (!\cref{fig:model}!, !\colorsquare{m_blue}!)
        patch_embeddings[~mask].reshape(N, -1, E),
        pos_embeddings[~mask].reshape(N, -1, E),
    )[-1]
    predictions = decoder(  # (N, S, E) (!\cref{fig:model}!, !\colorsquare{m_red}!)
        mask_embeddings.index_put([~mask], last_student_state.reshape(-1, E)),
        pos_embeddings,
    )[-1]
    predictions = regressor(predictions)  # (N, S, E)

    # optimization
    optimizer.zero_grad()
    loss = F.smooth_l1_loss(predictions[mask], targets[mask], beta=2)
    loss.backward()
    optimizer.step()

    # update teacher weights
    ema_update(student, teacher)
\end{pseudocode}
\end{figure*}

\end{document}